\documentclass{article}

\usepackage[preprint]{neurips_2026}

\usepackage[utf8]{inputenc} 
\usepackage[T1]{fontenc}    
\usepackage{url}            
\usepackage{booktabs}       
\usepackage{amsfonts}       
\usepackage{nicefrac}       
\usepackage{microtype}      
\usepackage{graphicx}       

\usepackage{xcolor}
\usepackage{colortbl}
\usepackage{multirow}
\definecolor{citecolor}{RGB}{4, 20, 110}
\usepackage[pagebackref=true,breaklinks=true,letterpaper=true,colorlinks,citecolor=citecolor,bookmarks=false]{hyperref}
\definecolor{demphcolor}{gray}{.5}

\newcommand{\stat}[2]{$#1$}

\title{Context-Aware Autoregressive Image Generation for Emerging Reasoning Properties}

\author{%
\begin{tabular}{c}
Jixuan Ying$^{1,2,*}$, Haoyu Liu$^{2,*,\dagger}$, Timing Yang$^{1}$, Tingyu Zhu$^{2}$\\
Predrag Neskovic$^{4}$, Anand Bhattad$^{1}$, Cihang Xie$^{3}$\\
Zeyu Zheng$^{2,\dagger}$, Alan Yuille$^{1}$, Feng Wang$^{1,\dagger}$\\[0.5ex]
{\normalfont $^{1}$Johns Hopkins University}\\
{\normalfont $^{2}$University of California, Berkeley\quad $^{3}$University of California, Santa Cruz}\\
{\normalfont $^{4}$Office of Naval Research}\\
{\normalfont \textsuperscript{*}Equal contribution\quad \textsuperscript{$\dagger$}Corresponding authors}\\
{\normalfont \texttt{haoyuliu@berkeley.edu}\quad \texttt{zyzheng@berkeley.edu}\quad \texttt{wangf3014@gmail.com}}
\end{tabular}
}

\begin{document}

\maketitle

\begin{abstract}
  Autoregressive image generation has recently emerged as a competitive visual generation paradigm, yet its generation order is often predefined or heuristic, such as next-patch prediction, random-order prediction, or one-step generation for specific scales. We observe that these strategies can cause models to overlook important structural dependencies in images, thereby limiting both structural robustness and reasoning capability. In this work, we propose a context-aware autoregressive generation framework, where the model dynamically determines the next token location based on global context and current local information. Rather than following a fixed or random order, our method allows the generation process to adapt to image structure and semantic dependencies. Extensive experiments show that context-aware generation significantly improves structural robustness across different image AR paradigms, including masked autoregressive and scale-wise autoregressive models. Beyond quantitative improvements, we further observe emerging reasoning properties: the model can better understand complex semantic instructions and generate images that more faithfully follow the given guidance. These findings suggest that generation order is a critical yet underexplored factor in image autoregressive modeling, and that context-aware decoding can unlock stronger reasoning-aware visual generation capabilities.
  Code is available at \url{https://github.com/Hpesojyjx/Saccades-AR}.
\end{abstract}

\section{Introduction}
\label{sec:intro}

Inspired by the success of next-token prediction in natural language processing, recent years have witnessed a growing interest in autoregressive (AR) image generation. Early efforts directly transferred the language modeling recipe to visual generation by flattening 2D images, pixels, or discrete visual tokens into 1D sequences and modeling them with standard autoregressive Transformers, following a raster or otherwise fixed token order~\citep{igpt,dalle,vqgan,parti,llamagen}. While conceptually simple and scalable, this direct transfer is not perfectly aligned with the spatial nature of images. Unlike text, where the generation order often follows a natural linguistic progression, images contain dense 2D structures and long-range spatial dependencies that do not naturally conform to a single left-to-right sequence. As a result, fixed-order AR generation can easily underutilize important structural cues, leading to visible artifacts such as row-wise or column-wise misalignment, distorted object layouts, and weakened global coherence. To alleviate these limitations, subsequent works explored more flexible generation schemes, including masked generative modeling that predicts randomly masked tokens or iteratively refines token sets~\citep{maskgit,muse,mage,mar,maskgen}, as well as scale-wise autoregressive generation that predicts visual tokens progressively from coarse to fine resolutions~\citep{var,star,infinity}, which reduce the rigidity of raster decoding and improve visual fidelity and structural stability.

\begin{figure}[t]
  \centering
  \includegraphics[width=\textwidth]{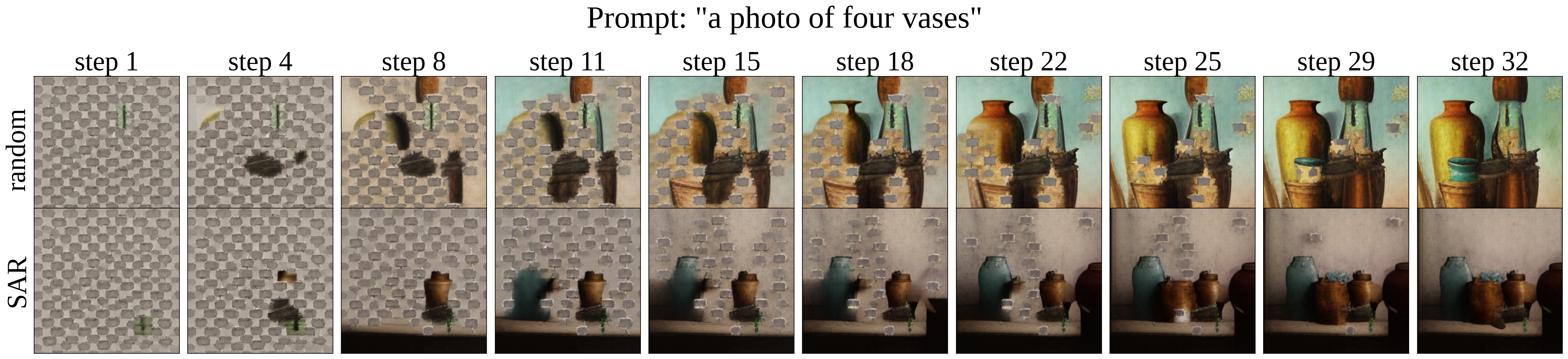}
  \caption{\textbf{\textit{Generation order shapes reasoning-like behaviors}} in autoregressive image generation. With random-order decoding (top), intermediate predictions remain scattered and incomplete, weakening feedback for counting (e.g., ``four vases") and often causing over-generation. With context-aware location selection (bottom), the same model and seed tend to complete objects more coherently (roughly one-by-one) and stop adding new ones once the count is met.}
  \label{fig:motivation}
\end{figure}

Nonetheless, although random-order and scale-wise generation alleviate the structural distortion, the reasoning logic behind patch-level generation remains largely overlooked. For example, \textbf{\textit{when humans draw an image patch by patch, they can usually decide where to draw next based on the existing context.}} This allows them to build a more coherent spatial organization and progressively reason over the partially completed content. In contrast, masked AR models generate patches in a random order, while scale-wise AR models generate all tokens within a given spatial scale in one step. Both strategies can easily miss an important intermediate reasoning process. We illustrate this issue with an intuitive example in Fig.~\ref{fig:motivation}, where the model is asked to generate \textit{four vases}. In Case 2, when the model is allowed to determine the next prediction location according to context, an emerging reasoning behavior can be observed. The model generates the vases in an organized manner, roughly one by one, and obtains self-feedback during the generation process. For instance, before step 22, it has already generated three vases. In the following steps, the model appears to recognize that one vase is still missing, and completes the fourth vase by step 25. After the required count is satisfied, the remaining steps are mainly used to fill in the background. In contrast, in Case 1, the random-order AR model selects prediction locations randomly at each step. As a result, most objects remain incomplete during the intermediate generation process, making it difficult for the model to receive reliable feedback about the object count. Consequently, it keeps generating new vase-like regions and fails to align with the required number.

Motivated by this observation, we propose \textbf{Saccade AR (SAR)}\footnote{We call our method \emph{Saccade} AR because it is inspired by saccadic eye movements in the human visual system, which actively choose where to fixate based on context when interpreting a visual scene.}, a simple yet surprisingly effective context-aware decoding strategy for AR image generation. In detail, at each decoding step, an AR model typically produces candidate predictions for all editable token positions, after which existing methods commit tokens according to a random, predefined, or scale-wise schedule. Instead, SAR ranks candidate token locations by their semantic relevance to the global context. Specifically, we measure the distance between each candidate token representation and the global conditioning signal, such as the class token in class-conditional generation or text embeddings in text-to-image generation, and prioritize locations from semantically close to semantically distant. In this way, SAR first generates regions that are strongly related to the global intent, and then fills in less critical details. SAR is training-free, lightweight, and broadly applicable. It can be plugged into pretrained AR image models at inference time without changing model parameters, and it introduces almost no additional computation because it only reorders which candidate predictions are committed. Moreover, SAR applies to different AR paradigms, including masked autoregressive models (MAR~\citep{mar}) and scale-wise autoregressive models (VAR~\citep{var}). Importantly, SAR does not change the model architecture or supervision; it changes the generation trajectory. Our central thesis is that \textbf{\textit{this trajectory matters}}: when the model generates in a context-aware order, it can better exploit intermediate visual feedback and exhibit stronger structural robustness and reasoning-related instruction following.

Extensive experiments validate this hypothesis. On text-to-image benchmarks such as GenEval~\citep{ghosh2023geneval} and DPG-Bench~\citep{hu2024ella}, SAR consistently improves pretrained AR generators including MaskGen~\citep{maskgen} and Infinity~\citep{infinity}, with especially clear gains on reasoning-related dimensions such as counting, relative positioning, and compositional instruction following. To further evaluate reasoning ability, we construct a long-prompt text-to-image benchmark involving complex combinations of object counts, spatial relations, and fine-grained attribute assignments. On this benchmark, SAR brings even larger improvements, supporting our claim that generation order is a critical factor behind the reasoning limitations of existing AR image models.

\section{Related Work}
\label{sec:related}

\paragraph{Autoregressive models in language and multimodal generation.} Recent advances in large language models (LLMs)~\citep{gpt3,llama1,llama2,team2023gemini,openai2024gpt4o,yang2024qwen2} have made autoregressive (AR) models central to modern generative modeling, due to their flexibility and wide use in predicting each next token from prior tokens. First developed for text, AR models are now also widely used for visual generation~\citep{igpt,vqgan,llamagen,tian2024visual,li2024autoregressive}, where each image is treated as a sequence of discrete tokens. The same idea has further led to unified vision--language models that handle both multimodal understanding and generation, such as Emu3~\citep{emu32024}, LaVIT~\citep{jin2023unified},and VILA-U~\citep{wu2024vila}. In these models, one shared AR backbone reads and writes mixed text and image tokens.

\paragraph{Autoregressive image generation with raster-scan token order.} Early research on AR visual modeling adopts language model architectures to generate images pixel by pixel~\citep{igpt}. With the help of image tokenizers~\citep{oord2017neural,vqgan}, later AR models produce images as discrete token sequences in a next-token manner, where each image is flattened into a 1D sequence and predicted by a decoder-only transformer, similar to language modeling. This pipeline has been scaled up to large-scale text-conditioned generation, such as DALL-E~\citep{dalle}, CogView~\citep{ding2021cogview,ding2022cogview2}, Make-A-Scene~\citep{gafni2022make}, Parti~\citep{parti}, LlamaGen~\citep{llamagen}, and Lumina-mGPT~\citep{liu2024lumina}, with parallel extensions to text-to-video models such as VideoPoet~\citep{kondratyuk2023videopoet}. However, these methods produce image tokens in raster-scan order, which does not fully use the 2D structure of natural images, leading to structural artifacts and weakened coherence under complex semantic instructions.

\paragraph{Improved generation orders for autoregressive image synthesis.}
Beyond raster-scan generation, recent works explore two
complementary directions. \emph{Mask-based methods} predict multiple tokens per step in a non-fixed order:
MaskGIT~\citep{maskgit} predicts groups of tokens jointly at each step, cutting the inference cost;
Muse~\citep{muse} extends this idea to text-to-image generation; MAGE~\citep{mage} unifies representation learning and synthesis under the same masked-token objective. MAR~\citep{li2024autoregressive} brings masked prediction to continuous-valued tokens via a small diffusion head, replacing the discrete codebook, and MaskGen~\citep{kim2025democratizing} compresses images into compact 1D tokens for masked text-to-image generation. \emph{Scale-wise methods} replace the spatial axis with a coarse-to-fine scale axis: VAR~\citep{tian2024visual} sums predictions across scales so the inference step count equals the
number of scales, with follow-ups in STAR~\citep{star} and Infinity~\citep{han2025infinity}. While both directions alleviate the rigidity of raster-scan, the resulting order is still either fixed \emph{a priori} (a predetermined scale schedule, as in VAR) or determined \emph{stochastically} (random or confidence-based unmasking, as in MaskGIT and MAR)---no existing method dynamically chooses the next step based on the semantic context already produced. This motivates our
\emph{context-aware} generation order, in which the model itself decides where to attend and generate next, analogous to the saccadic behaviour of the human visual system.

\section{Method}
\label{sec:method}

\begin{figure}[t]
    \centering
    \includegraphics[width=0.9\linewidth]{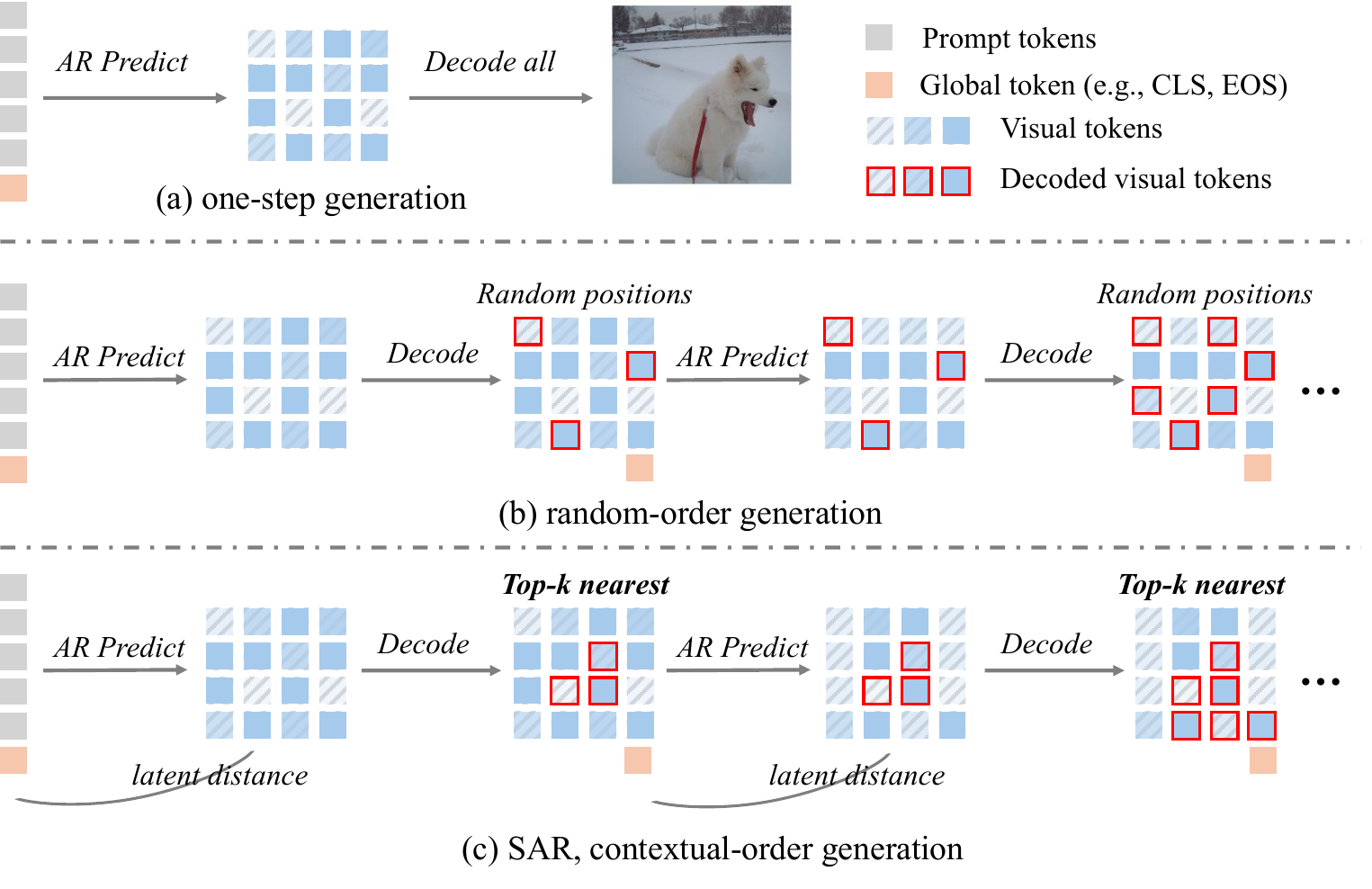}
    \caption{Comparison to one-step and random-order generation paradigms. Our SAR model decides the generation order based on each token's latent distance to a global token. \textbf{\textit{This is only a minimal change for image AR models, but we find it crucial for the AR model's reasoning behavior.}}}
    \label{fig:teaser_ar}
\end{figure}

\subsection{Image Autoregressive Models}
\label{sec:method:ar}

Modern image autoregressive models can be grouped into two families that differ in the unit of generation. \emph{Patch-wise} models tokenize the image into a flat sequence of spatial (or 1D-compressed) latent tokens and predict them one or a few at a time. \emph{Scale-wise} models predict the image at progressively finer resolutions, generating a coarse-to-fine pyramid of tokens. We describe the patch-wise scale-wise formulations below, and then explain how SAR adapt with them.

\paragraph{Patch-wise masked autoregressive models.}
We adopt the masked-autoregressive (MAR) formulation~\citep{li2024autoregressive}, encoding an image $x$ into $N$ continuous latent tokens $z = \{z_i\}_{i=1}^{N}$ and factorizing the conditional distribution over a generation order $\pi$:
\begin{equation}
p_\theta(z \mid c) \;=\; \prod_{t=1}^{T} p_\theta\!\left(z_{S_t} \,\middle|\, z_{S_{<t}},\, c\right),
\label{eq:mar}
\end{equation}
Here $c$ is a conditioning signal, and $\{S_t\}_{t=1}^{T}$ partitions token indices according to $\pi$. At step $t$, the model predicts $S_t$ in parallel given $z_{S_{<t}}$, unmasking $|S_t|$ tokens with the cosine schedule above. Each per-token conditional is implemented by an MAE-style transformer decoder with a lightweight denoising-diffusion head, and we apply classifier-free guidance (CFG) with a linear schedule across AR steps at inference.

\paragraph{Scale-wise autoregressive models.}
We also consider scale-wise AR as in VAR~\citep{tian2024visual} and Infinity~\citep{han2025infinity}, which generates a coarse-to-fine pyramid $z=\{z^{(k)}\}_{k=1}^{K}$ (with $N_k$ tokens at scale $k$):
\begin{equation}
p_\theta(z \mid c)
\;=\;
\prod_{k=1}^{K}
p_\theta\!\left(z^{(k)} \,\middle|\, z^{(<k)},\, c\right),
\label{eq:scale-ar}
\end{equation}
At scale $k$, the model predicts the next finer tokens conditioned on coarser scales $z^{(<k)}$ and $c$, progressively refining details. Instantiations differ across models: VAR uses discrete next-scale tokens, while Infinity predicts bits with a multi-scale residual tokenizer.

\subsection{Saccade Autoregressive Generation}
\label{sec:method:sar}

The order $\pi$ in Eq.~\ref{eq:mar} is typically fixed (e.g., raster/zigzag) or sampled uniformly at random~\citep{li2024autoregressive}. Neither leverages image spatial semantics. Inspired by human saccades, we instead choose the order adaptively at inference time, generating the tokens most relevant to the conditioning first, with saliency decided by the model rather than a hand-designed scan. Let $h_i^{(t)} \in \mathbb{R}^d$ denote the model-specific candidate representation used for ranking at masked position $i \in \mathcal{M}_t$ and AR step $t$. Let $\mathcal{A}^{(t)} \subset \mathbb{R}^d$ be conditioning \emph{anchors} in the same embedding space: the conditioning tokens visible before any image token is generated (class tokens for class-conditional, prompt tokens for text-conditional). Model-specific details are in Sec.~\ref{sec:exp}. Because the generation pipeline produces candidate representations for \emph{all} masked positions at each step, we can rank positions on the fly. We score each still-masked position by its average cosine similarity to anchors,
\begin{equation}
s_i^{(t)} \;=\; \frac{1}{|\mathcal{A}^{(t)}|}\sum_{a \in \mathcal{A}^{(t)}} \frac{\langle h_i^{(t)},\, a \rangle}{\|h_i^{(t)}\|\,\|a\|},
\qquad i \in \mathcal{M}_t,
\label{eq:sar-score}
\end{equation}
then unmask the top-$|S_t|$ positions (same cosine schedule as Eq.~\ref{eq:mar}). Recomputing $h^{(t)}$ each step yields an adaptive, context-dependent order. SAR changes only inference-time ranking and adds a single similarity computation per (masked token, anchor, step).

\paragraph{MAR feature selection.}
For MAR, position selection is performed after token-wise diffusion denoising, rather than directly on the deterministic masked-prediction features. At each step, the model first predicts candidates for unresolved positions from the current visible context, and then samples denoised candidate tokens before SAR selects which positions to commit. Diffusion sampling introduces seed-dependent variation, so the first selected position need not be fixed for a given class condition. After each commitment, the updated partial image conditions the next round of candidate generation and ranking.

\paragraph{SAR for scale-wise generation.}
Scale-wise AR (Eq.~\ref{eq:scale-ar}) generates all $N_k$ tokens at a scale in one shot, so there is no intra-scale order. We introduce an order by decoding only the final scale $K$ over $M{>}1$ steps (scales $1\ldots K{-}1$ unchanged), since the last scale has the highest spatial resolution and contains most fine details. We split the $N_K$ final-scale positions into $M$ equal groups and commit one group per step. At step $m$, committed positions use their decoded embeddings and the rest keep the coarse features from scales $1\ldots K{-}1$; a non-causal pass produces fresh hidden states $h_i^{(m)}$ for all uncommitted positions. We rank positions by squared L2 distance to an anchor $a$,
\begin{equation}
s_i^{(m)} \;=\; \bigl\|h_i^{(m)} - a\bigr\|_2^2,
\label{eq:sar-scale}
\end{equation}
and commit the $N_K/M$ positions with smallest scores. We reuse the model's existing conditioning embedding as $a$ (class embedding $c_{\mathrm{class}}$ or pooled text embedding $c_{\mathrm{pool}}$), computed once and fixed across steps, so this adds essentially no extra overhead.

Overall, scale-wise SAR is an inference-time re-ranking on top of $M$-step last-scale decoding; it requires no parameter updates and adds one similarity per uncommitted position per step. We instantiate it on VAR~\citep{tian2024visual} (class-conditional; Appendix~\ref{app:class}) and Infinity~\citep{han2025infinity} (text-to-image; Sec.~\ref{sec:exp:t2i}). Table~\ref{tab:imagenet-fid} compares inference-time orders on shared checkpoints. Separately, Table~\ref{tab:var_sar_retrain} studies VAR retrained using SAR ordering at the last scale, followed by evaluation under different inference orders.

\section{Experiments}
\label{sec:exp}

In this section, we evaluate SAR across both patch-wise and scale-wise AR. Sec.~\ref{sec:exp:qual} qualitatively visualizes the intermediate decoded image at every AR step under random vs.\ SAR; Sec.~\ref{sec:exp:t2i} reports our main text-to-image results on MaskGen and Infinity, covering GenEval, DPG-Bench, MJHQ-30K FID, and a complex long-prompt reasoning benchmark we construct (Sec.~\ref{sec:exp:long_prompt}); Sec.~\ref{sec:exp:ablation} presents ablations on alternative orders and on the VAR retraining. We also provide class-conditional ImageNet results on MAR and VAR in Appendix~\ref{app:class}.

\subsection{Qualitative Analysis}
\label{sec:exp:qual}

Before the quantitative results we build intuition for \emph{how} SAR helps by visualizing the intermediate decoded image at every AR step under random vs.\ SAR, with the checkpoint and prompt held fixed. We use one representative model per AR family: Infinity-8B (scale-wise) and MaskGen-XL (patch-wise). Implementation details are provided in Sec.~\ref{sec:exp:t2i}. The two realize SAR's improvement through \emph{different} reasoning behaviors that converge on the same effect.

\begin{figure}[htbp]
\centering
\includegraphics[width=0.95\linewidth]{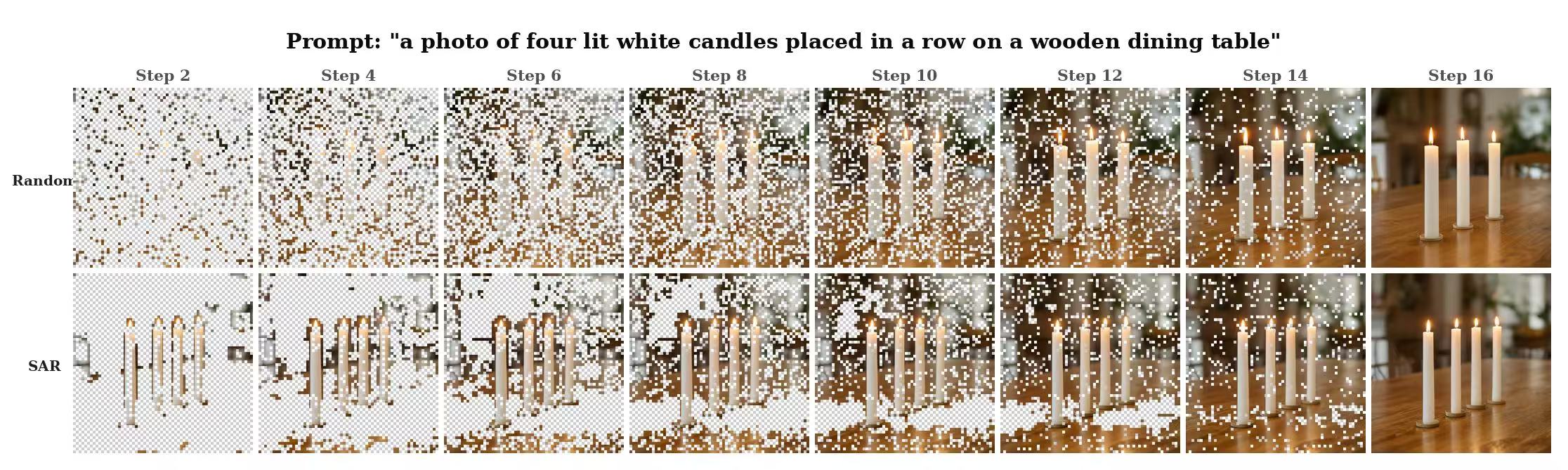}\\[2pt]
\includegraphics[width=0.95\linewidth]{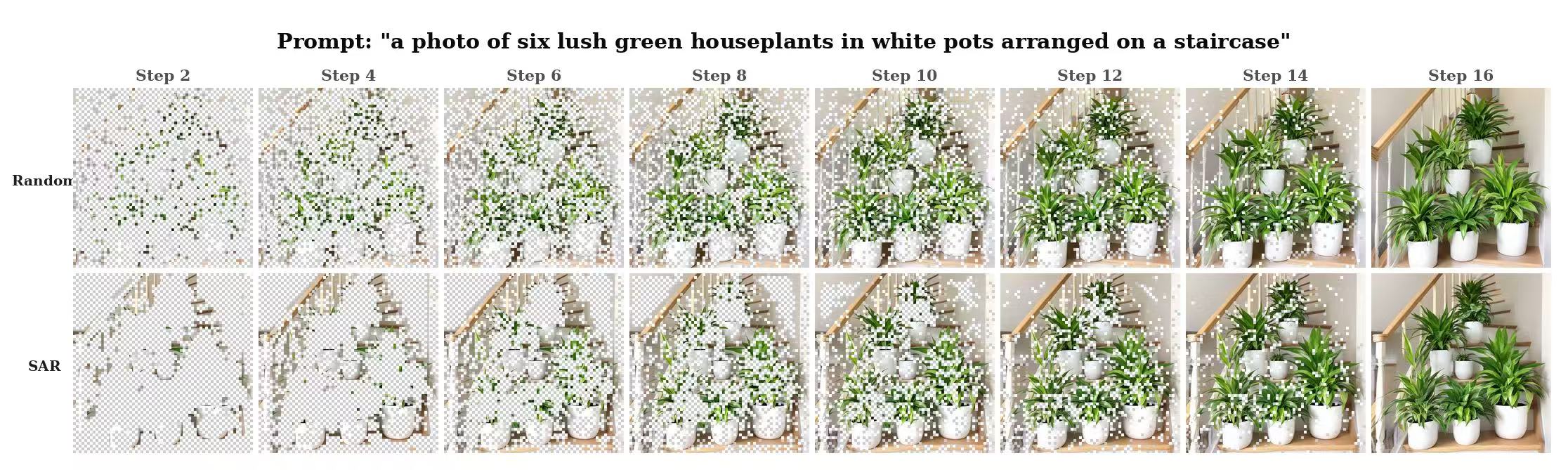}\\[2pt]
\includegraphics[width=0.95\linewidth]{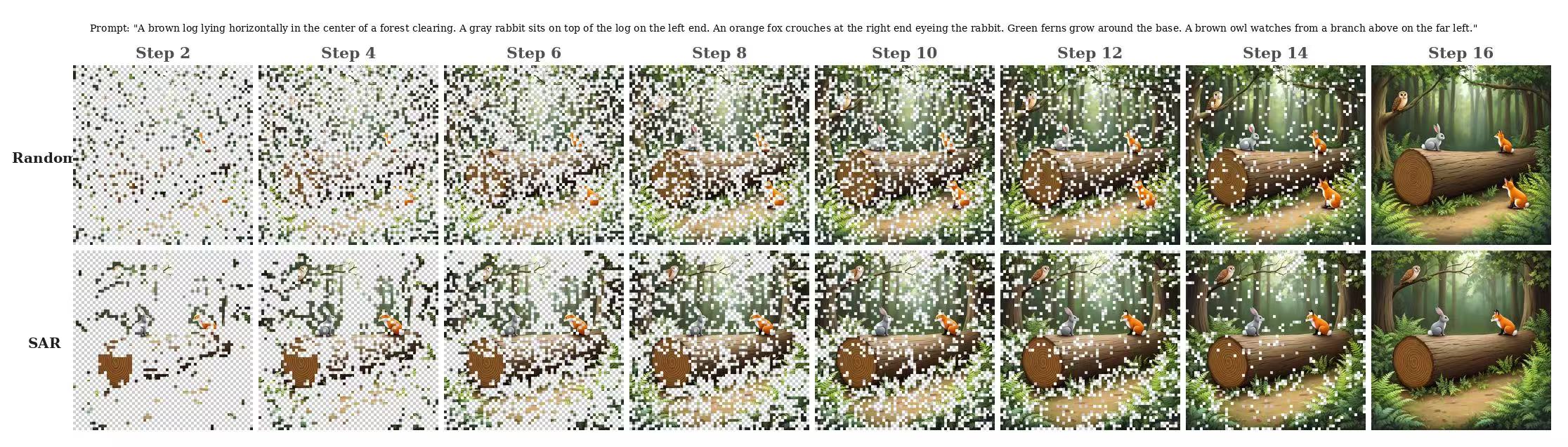}\\[2pt]
\includegraphics[width=0.95\linewidth]{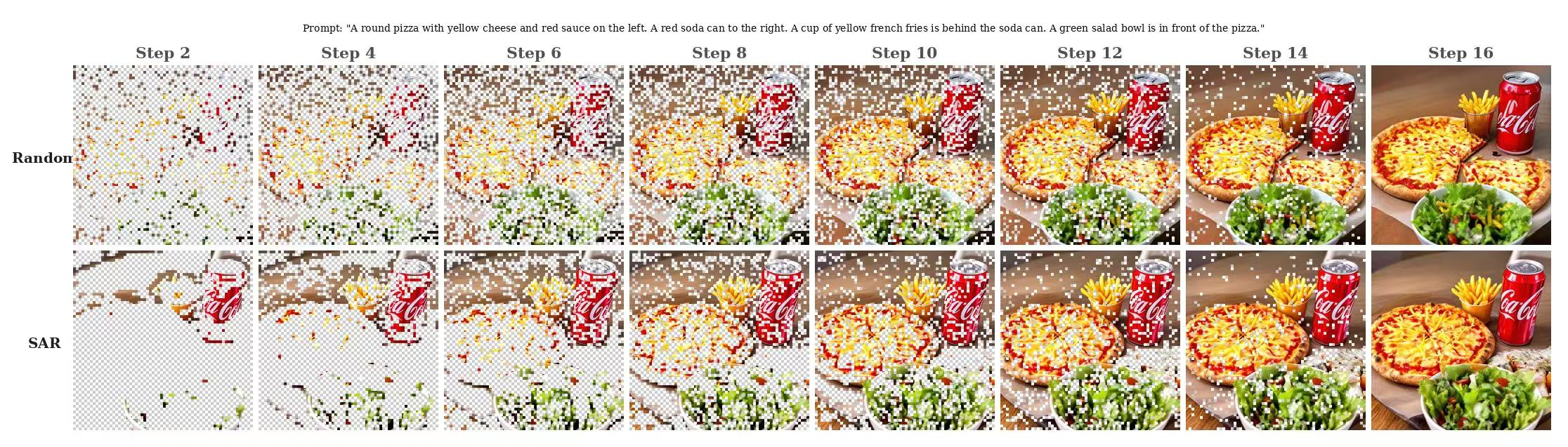}
\caption{Infinity-8B: \emph{outline-to-detail} reasoning under SAR. In each strip, the top row is random ordering and the bottom row is SAR; columns are AR steps. Within the first few SAR steps the global outline already encodes the correct count (four candles, six houseplants), the correct object set with no hallucinated extras (fox+rabbit), and the correct global geometry (circular pizza). Subsequent steps fill in textures without changing those structural decisions. Random ordering instead grows partial textures everywhere and frequently fails to recover count, identity, or shape by the final step.}
\label{fig:viz-infinity}
\end{figure}

\paragraph{Infinity: outline first, details later.}
On Infinity-8B (Fig.~\ref{fig:viz-infinity}), SAR commits to the prompt-correct \emph{global outline} (count, object identity, and overall geometry) within the first few last-scale steps, and uses subsequent steps only to fill in textures. Random ordering scatters partial textures before any outline is established and often fails to recover those structural decisions by the final step.

\begin{figure}[htbp]
\centering
\includegraphics[width=0.95\linewidth]{figs/order_illustrate.png}\\[2pt]
\includegraphics[width=0.95\linewidth]{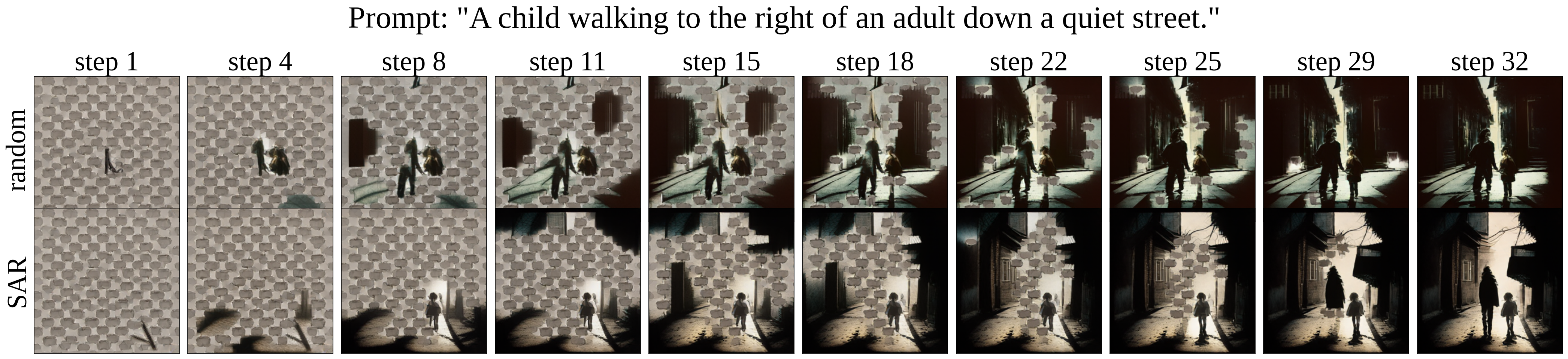}\\[2pt]
\includegraphics[width=0.95\linewidth]{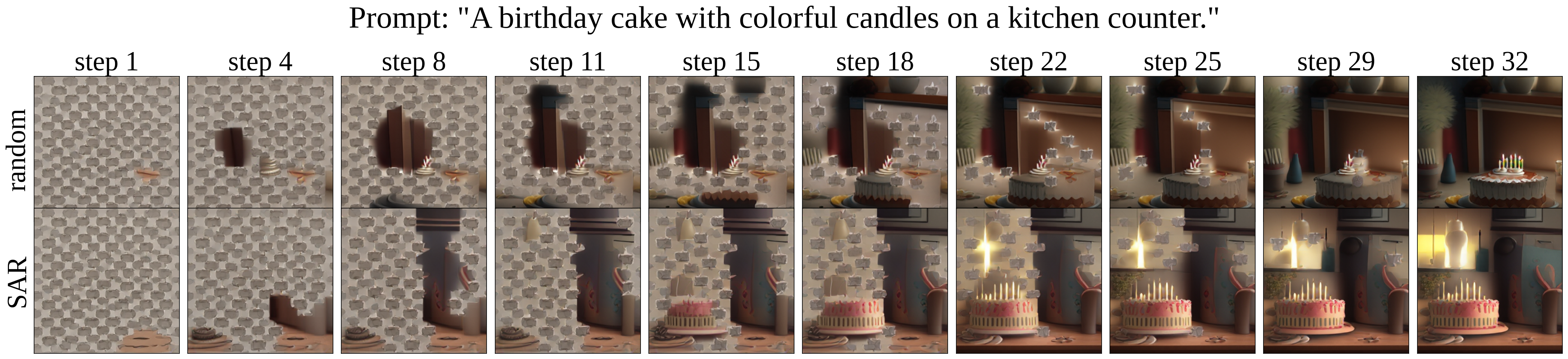}
\caption{MaskGen-XL: \emph{object-by-object} reasoning under SAR. Top to bottom: ``four vases'', ``a child walking to the right of an adult down a quiet street'', and ``a birthday cake with colorful candles on a kitchen counter''. SAR completes one object before starting the next (counting), commits one entity before introducing the other (spatial relation), and follows scene $\to$ main object $\to$ parts (compositional hierarchy); random ordering grows everything concurrently and only resolves identity and layout in the last few steps.}
\label{fig:viz-maskgen}
\end{figure}

\paragraph{MaskGen: one object at a time.}
On MaskGen-XL (Fig.~\ref{fig:viz-maskgen}), SAR instead commits one full object at a time, moving on only once the previous object is complete; counting, spatial relations, and compositional hierarchy all fall out of this sequential commitment. Random ordering grows all parts concurrently and resolves identity and layout only in the last few steps.

The two families realize SAR through complementary mechanisms (\emph{outline-to-detail} vs.\ \emph{object-by-object}), but in both cases SAR's prompt-grounded ordering turns AR decoding into a sequence of legible, prompt-aligned commitments rather than a soup of half-formed concepts. This implicit visual planning emerges \emph{for free}, and the quantitative results across all $8$ models in Sec.~\ref{sec:exp:t2i} and Appendix~\ref{app:class} confirm that this behavioral change translates into consistent benchmark gains.

\subsection{Text-to-image generation}
\label{sec:exp:t2i}

\paragraph{Setup.}
We evaluate text-to-image SAR on two AR families that span the dominant design choices in the literature: \textbf{MaskGen}~\citep{kim2025democratizing} (patch-wise; MaskGen-L and MaskGen-XL) and \textbf{Infinity}~\citep{han2025infinity} (scale-wise; Infinity-2B and Infinity-8B). MaskGen takes the SAR anchor to be the singleton $\{W\, c_{\mathrm{pool}}\}$, with $W$ the same linear projection MaskGen already uses to inject pooled text conditioning; Infinity takes the SAR anchor to be $a = c_{\mathrm{pool}}$, the attentive-pooled text embedding already consumed by AdaLN. We use $T{=}32$ AR steps for MaskGen and $M{=}16$ equal groups at the last scale for Infinity. All other implementation details follow the original recipes and are listed in Appendix~\ref{app:impl}.

\paragraph{Evaluation.}
We use three complementary benchmarks: GenEval~\citep{ghosh2023geneval} for compositional capability, with $4$ samples per prompt and the prompt-level pass criterion; DPG-Bench~\citep{hu2024ella} for fine-grained prompt alignment, scored by the mPLUG-VQA evaluator on a $2{\times}2$ image grid; and MJHQ-30K FID~\citep{li2024playground} for distribution-level aesthetic quality on a 30K-prompt subset. Following the original Infinity protocol, GenEval is run with prompt rewriting on Infinity; all other settings use raw prompts. All three are averaged over $3$ seeds.

\paragraph{Results.}
Tables~\ref{tab:geneval} and~\ref{tab:dpg} show that the inference-only SAR change improves compositional capability at every scale across both paradigms. On GenEval, the Overall pass rate rises consistently, with the largest gains on sub-categories that require global consistency such as \emph{Counting}, \emph{Colors}, and \emph{Position}. On DPG-Bench, the per-category L1 breakdown is mixed but the aggregate Overall score also improves at every scale. On MJHQ-30K FID (Table~\ref{tab:mjhq-fid}), SAR is statistically tied with random at every scale, indicating that the compositional gains do not come at the price of distribution-level aesthetic quality. Together these results demonstrate that a single inference-time intervention generalizes across both patch-wise and scale-wise text-to-image AR. For the scale-wise setting, Appendix~\ref{app:var-retrain} additionally studies SAR-order last-scale retraining of VAR and compares inference orders on the resulting checkpoints.

\subsubsection{Complex long-prompt benchmark}
\label{sec:exp:long_prompt}

To complement GenEval and DPG-Bench, we further build a complex long-prompt benchmark to better evaluate reasoning-intensive instruction following. The benchmark contains 2,000 prompts with dense compositional constraints, including object counts, spatial relations, attribute binding, object existence, compositional ordering, and view consistency. A typical prompt is: ``A vertical stack of four different items: a wooden donut at the bottom, a metallic sphere in the middle, and a miniature cactus on the very top. This stack is placed inside a hollow red ring. To the front-left of the stack is a spilled glass of milk, and to the back-right is a closed leather book. Cinematic top-down 45-degree angle view.''

For evaluation, we generate images from each prompt using different AR models and decoding orders. We then use Gemini 2.5 Pro~\citep{gemini25} as a VLM judge. For each image, we construct 10 question-answer pairs from the prompt, with ground-truth answers directly extracted from the prompt constraints. Gemini is asked to score the match between the generated image and each question on a 1--5 scale. We report the average scores under six reasoning-related categories.

\begin{table}[t]
\centering
\small
\setlength{\tabcolsep}{3.5pt}
\caption{
Results on our complex long-prompt benchmark. Scores are judged by Gemini 2.5 Pro on a 1--5 scale.
}
\label{tab:long_prompt_benchmark}
\begin{tabular}{llccccccc}
\toprule
Model & Order
& Cnt.
& Spat.
& Attr.
& Exist.
& Ord.
& View
& Avg. \\
\midrule
\multirow{2}{*}{MaskGen-XL}
& random & 2.55 & 2.24 & 2.39 & 2.51 & 2.32 & 2.51 & 2.42 \\
& SAR    & \textbf{3.08} & \textbf{2.78} & \textbf{2.94} & \textbf{3.01} & \textbf{2.82} & \textbf{2.89} & \textbf{2.92} \\
\midrule
\multirow{2}{*}{Infinity-2B}
& random & 2.76 & 2.43 & 2.64 & 2.72 & 2.49 & 2.68 & 2.62 \\
& SAR    & \textbf{3.27} & \textbf{2.98} & \textbf{3.16} & \textbf{3.22} & \textbf{2.99} & \textbf{3.10} & \textbf{3.12} \\
\midrule
\multirow{2}{*}{Infinity-8B}
& random & 3.09 & 2.74 & 2.94 & 3.02 & 2.82 & 2.91 & 2.92 \\
& SAR    & \textbf{3.61} & \textbf{3.31} & \textbf{3.45} & \textbf{3.49} & \textbf{3.30} & \textbf{3.36} & \textbf{3.42} \\
\bottomrule
\end{tabular}
\end{table}

Table~\ref{tab:long_prompt_benchmark} shows that SAR consistently improves complex instruction following across all three evaluated AR models, raising the average score by roughly $+0.5$ on each. The largest gains tend to appear on categories that require global consistency (counting, spatial relation, and compositional ordering) where random-order decoding fails to maintain reliable intermediate feedback about how many objects have already been generated or where they should be placed. This supports our hypothesis that generation order is not merely a low-level sampling detail but directly affects whether an AR image model can use partially generated content as feedback and reason over the remaining constraints during generation.

\begin{table}[t]
\centering
\caption{GenEval pass rate (\%, $\uparrow$) for both AR paradigms. Best per (model, metric) pair in bold.}
\label{tab:geneval}
\small
\setlength{\tabcolsep}{3pt}
\renewcommand{\arraystretch}{1.15}
\begin{tabular}{llccccccc}
\toprule
Model & Method & S.Obj & T.Obj & Count & Colors & Pos. & C.Attr & Overall \\
\midrule
\multirow{2}{*}{MaskGen-L}   & Random & $\mathbf{100.00}$            & $54.55$                     & $40.00$                     & $85.11$                     & $18.00$                     & $26.00$                     & $53.94$                     \\
                             & SAR    & $\mathbf{100.00}$            & $\mathbf{56.57}$            & $\mathbf{50.00}$            & $\mathbf{89.36}$            & $\mathbf{20.00}$            & $\mathbf{29.00}$            & $\mathbf{57.49}$            \\
\midrule
\multirow{2}{*}{MaskGen-XL}  & Random & $\mathbf{98.75}$             & $58.59$                     & $43.75$                     & $87.23$                     & $21.00$                     & $28.00$                     & $56.22$                     \\
                             & SAR    & $\mathbf{98.75}$             & $\mathbf{64.65}$            & $\mathbf{46.25}$            & $\mathbf{88.30}$            & $\mathbf{25.00}$            & $\mathbf{37.00}$            & $\mathbf{59.99}$            \\
\midrule
\multirow{2}{*}{Infinity-2B} & Random & \stat{\mathbf{100.00}}{0.00} & \stat{82.62}{2.01}          & \stat{64.38}{0.51}          & \stat{82.77}{1.28}          & \stat{\mathbf{42.10}}{1.23} & \stat{58.88}{1.61}          & \stat{71.79}{0.46}          \\
                             & SAR    & \stat{\mathbf{100.00}}{0.00} & \stat{\mathbf{83.34}}{1.50} & \stat{\mathbf{65.27}}{2.44} & \stat{\mathbf{85.42}}{1.38} & \stat{41.00}{2.16}          & \stat{\mathbf{60.38}}{0.98} & \stat{\mathbf{72.57}}{0.65} \\
\midrule
\multirow{2}{*}{Infinity-8B} & Random & \stat{\mathbf{100.00}}{0.00} & \stat{\mathbf{92.04}}{0.68} & \stat{68.69}{3.45}          & \stat{86.03}{1.57}          & \stat{\mathbf{66.76}}{0.33} & \stat{\mathbf{70.24}}{2.07} & \stat{80.63}{0.60}          \\
                             & SAR    & \stat{\mathbf{100.00}}{0.00} & \stat{91.76}{0.64}          & \stat{\mathbf{73.21}}{4.13} & \stat{\mathbf{86.74}}{1.18} & \stat{\mathbf{66.76}}{0.33} & \stat{69.57}{1.23}          & \stat{\mathbf{81.34}}{0.81} \\
\bottomrule
\end{tabular}
\end{table}

\begin{table}[t]
\centering
\caption{DPG-Bench L1 category breakdown (\%, $\uparrow$) for both AR paradigms. Best per (model, metric) pair in bold.}
\label{tab:dpg}
\small
\setlength{\tabcolsep}{4pt}
\renewcommand{\arraystretch}{1.15}
\begin{tabular}{llcccccc}
\toprule
Model & Method & Entity & Attribute & Relation & Global & Other & Overall \\
\midrule
\multirow{2}{*}{MaskGen-L}   & Random & \stat{\mathbf{81.82}}{0.21} & \stat{80.04}{0.50}          & \stat{80.91}{0.62}          & \stat{76.12}{1.56}          & \stat{81.08}{0.96}          & \stat{70.11}{0.06}          \\
                             & SAR    & \stat{78.30}{0.78}          & \stat{\mathbf{80.84}}{0.60} & \stat{\mathbf{82.91}}{1.49} & \stat{\mathbf{82.75}}{2.57} & \stat{\mathbf{81.70}}{1.74} & \stat{\mathbf{70.51}}{0.19} \\
\midrule
\multirow{2}{*}{MaskGen-XL}  & Random & \stat{79.33}{0.87}          & \stat{\mathbf{80.98}}{0.63} & \stat{81.63}{1.81}          & \stat{80.81}{1.17}          & \stat{80.52}{1.96}          & \stat{71.00}{0.08}          \\
                             & SAR    & \stat{\mathbf{81.53}}{0.58} & \stat{79.82}{0.69}          & \stat{\mathbf{82.83}}{1.20} & \stat{\mathbf{82.81}}{0.99} & \stat{\mathbf{81.08}}{0.48} & \stat{\mathbf{71.57}}{0.24} \\
\midrule
\multirow{2}{*}{Infinity-2B} & Random & \stat{89.14}{0.04}          & \stat{86.19}{0.25}          & \stat{92.50}{0.08}          & \stat{\mathbf{83.18}}{0.62} & \stat{82.67}{2.29}          & \stat{82.71}{0.23}          \\
                             & SAR    & \stat{\mathbf{89.24}}{0.15} & \stat{\mathbf{86.23}}{0.38} & \stat{\mathbf{92.51}}{0.10} & \stat{82.88}{0.29}          & \stat{\mathbf{82.93}}{2.22} & \stat{\mathbf{82.76}}{0.22} \\
\midrule
\multirow{2}{*}{Infinity-8B} & Random & \stat{91.38}{0.21}          & \stat{88.75}{0.02}          & \stat{94.24}{0.05}          & \stat{83.99}{0.57}          & \stat{\mathbf{86.80}}{0.86} & \stat{86.09}{0.07}          \\
                             & SAR    & \stat{\mathbf{91.45}}{0.14} & \stat{\mathbf{88.87}}{0.05} & \stat{\mathbf{94.25}}{0.15} & \stat{\mathbf{85.01}}{0.62} & \stat{86.67}{1.61}          & \stat{\mathbf{86.27}}{0.08} \\
\bottomrule
\end{tabular}
\end{table}

\begin{table}[t]
\centering
\caption{MJHQ-30K FID ($\downarrow$) for both AR paradigms. Across all four scales, SAR is statistically tied with the random-order baseline, indicating that context-aware ordering does not change the distribution-level aesthetic quality of the generated images.}
\label{tab:mjhq-fid}
\small
\setlength{\tabcolsep}{6pt}
\begin{tabular}{lcccc}
\toprule
 & \multicolumn{2}{c}{Patch-wise (MaskGen)} & \multicolumn{2}{c}{Scale-wise (Infinity)} \\
 \cmidrule(lr){2-3} \cmidrule(lr){4-5}
Order  & MaskGen-L & MaskGen-XL & Infinity-2B & Infinity-8B \\
\midrule
Random & \stat{6.79}{0.04} & \stat{6.75}{0.04} & \stat{10.867}{}    & \stat{9.332}{} \\
SAR    & \stat{6.79}{0.03} & \stat{6.74}{0.02} & \stat{10.970}{}    & \stat{9.317}{} \\
\bottomrule
\end{tabular}
\end{table}

\subsection{Ablation Study}
\label{sec:exp:ablation}

To check that the gains of SAR are not a trivial side-effect of merely \emph{abandoning} the random baseline, we compare it against two deterministic alternatives: \emph{Sequential}, which generates tokens one by one in their natural index order, and \emph{Inside-out}, which starts from the middle token and expands outward symmetrically. We evaluate them on one representative model from each AR family: MaskGen-XL (patch-wise) and Infinity-8B (scale-wise). Table~\ref{tab:order-ablation} reports GenEval and DPG-Bench Overall scores under all four orders. SAR consistently outperforms all three alternative orderings on both benchmarks and on both models, indicating that what matters is conditioning the choice of next position on the prompt itself, rather than simply imposing a structural pattern.

\begin{table}[htbp]
\centering
\caption{Ablation on the inference-time generation order, on one representative model from each AR family. ``Sequential'' generates tokens one by one in their natural index order; ``Inside-out'' starts from the middle token and expands outward. Higher is better for all columns; best per column in bold.}
\label{tab:order-ablation}
\small
\setlength{\tabcolsep}{6pt}
\begin{tabular}{lcccc}
\toprule
 & \multicolumn{2}{c}{MaskGen-XL} & \multicolumn{2}{c}{Infinity-8B} \\
\cmidrule(lr){2-3} \cmidrule(lr){4-5}
Order & GenEval ($\uparrow$) & DPG-Bench ($\uparrow$) & GenEval ($\uparrow$) & DPG-Bench ($\uparrow$) \\
\midrule
Random          & 0.55          & 71.00          & 0.804          & 86.05          \\
Sequential      & 0.57          & 71.00          & 0.799          & 86.01          \\
Inside-out      & 0.55          & 71.30          & 0.808          & 86.25          \\
SAR (ours)      & \textbf{0.58} & \textbf{71.57} & \textbf{0.815} & \textbf{86.36} \\
\bottomrule
\end{tabular}
\end{table}

\paragraph{Controlled comparison with confidence scheduling.}
MaskGIT~\citep{maskgit} selects positions using prediction confidence, whereas SAR explicitly ranks candidate representations by their alignment with the conditioning anchor. Both rules can depend on the current context; the distinction is the scoring criterion, not dynamic selection alone. We replace the confidence selector in a pretrained MaskGIT-Base checkpoint with SAR, keeping the checkpoint, 12-step cosine schedule, classifier-free guidance setting, and 50K class-conditional ImageNet $256\times256$ sample count fixed, without retraining. Table~\ref{tab:maskgit-confidence} shows lower FID and higher recall under SAR, isolating the benefit of the selection criterion.

\begin{table}[!t]
\centering
\caption{Controlled MaskGIT-Base comparison: only the position-selection criterion changes.}
\label{tab:maskgit-confidence}
\begin{tabular}{lcc}
\toprule
Selection rule & FID $\downarrow$ & Recall $\uparrow$ \\
\midrule
Confidence (original) & 6.48 & 0.41 \\
SAR & \textbf{5.95} & \textbf{0.43} \\
\bottomrule
\end{tabular}
\end{table}

\paragraph{Inference cost and matched-budget comparisons.}
We measure VAR end-to-end generation latency on one NVIDIA H100 80GB GPU with BF16 and batch size $1$. After $20$ warm-up generations, we time $1{,}000$ ImageNet images, including VAE decoding but excluding model loading and image saving. Random and SAR use the same retrained checkpoint and identical four-step last-scale decoding, differing only in position selection. Table~\ref{tab:var-latency} in Appendix~\ref{app:latency} reports mean latency and population standard deviation alongside generation quality. Across all four VAR sizes, Random and SAR have identical mean latency at the reported precision, although their timing variability can differ. This indicates no measurable additional overhead from SAR's ranking in this setup, rather than zero computational cost. The increase relative to one-step decoding comes from the shared multi-step last-scale procedure; the matched four-step comparison isolates the effect of the ordering rule.

\section{Conclusion}
\label{sec:conclusion}

We introduced \textbf{SAR}, a context-aware generation order that, at every AR step, ranks the still-uncommitted positions by similarity to a prompt-derived anchor (the conditioning embedding the base model already maintains) and commits the most relevant ones first; SAR introduces no new parameters and runs purely at inference. Across $8$ checkpoints spanning both AR paradigms (MAR, MaskGen, VAR, Infinity) and four standard benchmarks plus our complex long-prompt benchmark, SAR consistently improves prompt alignment and compositional reasoning while preserving distribution-level aesthetic quality, and qualitatively elicits two complementary reasoning behaviors: outline-first decoding in scale-wise models and object-by-object decoding in patch-wise models. The fact that this implicit visual planning emerges for free from the model's own representations suggests that generation order is a critical and underexplored axis of AR image modeling, with richer anchors and combinations with aesthetic-aware criteria as natural directions for further gains.

\begin{ack}
Authors from Johns Hopkins University are supported by ONR grants N000142512132 and N000142412696.
\end{ack}

\newpage
\bibliographystyle{ieeenat_fullname}
\bibliography{references}

\newpage
\appendix

\section{Implementation Details}
\label{app:impl}

Code for the VAR and Infinity implementations is publicly available at
\url{https://github.com/Hpesojyjx/Saccades-AR}.

\subsection{Text-to-image generation (Sec.~\ref{sec:exp:t2i})}

\paragraph{MaskGen.}
We use the public MaskGen-L and MaskGen-XL checkpoints~\citep{kim2025democratizing}, both built on the KL variant of the TATiTok 1D image tokenizer that compresses an image into $N{=}32$ continuous latent tokens. A frozen CLIP ViT-L/14-336 text encoder produces a per-token text feature sequence and a pooled embedding $c_{\mathrm{pool}}$; the linear projection $W$ used to form the SAR anchor is the same one MaskGen already uses to inject pooled text conditioning into its decoder. We sample with $T{=}32$ AR steps, CFG scale $3.0$ with a linear schedule, and an aesthetic micro-conditioning score of $6.5$, matching the original MaskGen recipe.

\paragraph{Infinity.}
We use Infinity-2B and Infinity-8B~\citep{han2025infinity}, both built on the BSQ-VAE tokenizer with $K{=}13$ coarse-to-fine scales at $1024{\times}1024$ resolution; the final scale alone contains $4096$ spatial positions and accounts for the bulk of the token budget. A frozen FLAN-T5-XL text encoder produces a per-token feature sequence consumed by cross-attention and a learned attentive-pooled embedding $c_{\mathrm{pool}}$ used for AdaLN modulation. We run multi-step inference at the final scale only, with $M{=}16$ equal groups, leaving scales $1$ through $K{-}1$ unchanged. CFG is $4.0$ for Infinity-2B and $3.0$ for Infinity-8B with sampling temperature $\tau{=}1$, matching the original Infinity recipe.

\subsection{Class-conditional generation (Appendix~\ref{app:class})}

\paragraph{MAR.}
We use MAR-Base ($208$M) and MAR-Large ($479$M)~\citep{li2024autoregressive}. The image is encoded by a KL-regularized 2D VAE into $N{=}256$ continuous spatial tokens of dimension $16$, and the class label is injected as a buffer of class-token embeddings prepended to the decoder input. We sample with $T{=}256$ AR steps, $100$ diffusion sampling steps inside the per-token diffusion head, and CFG with a linear schedule (CFG scale $2.9$ for MAR-Base, $3.0$ for MAR-Large), all matching the original MAR recipe.

\paragraph{VAR.}
We use VAR-$d16$ ($310$M) and VAR-$d20$ ($600$M)~\citep{tian2024visual}. Each image is encoded by a multi-scale residual VQ tokenizer into a coarse-to-fine pyramid of $K{=}10$ token maps with final resolution $16{\times}16$ at $256{\times}256$ output, and the class label is injected as a class embedding $c_{\mathrm{class}}{=}\mathrm{ClassEmb}(y)$ driving AdaLN modulation. We run last-scale multi-step inference with $M{=}4$ equal groups, leaving scales $1$ through $K{-}1$ unchanged, and adopt the CFG schedule and sampling temperature of the original VAR recipe.

\section{Class-conditional Generation Results}
\label{app:class}

\paragraph{Setup.}
We evaluate class-conditional SAR on two AR families: \textbf{MAR}~\citep{li2024autoregressive} (patch-wise; MAR-Base and MAR-Large) and \textbf{VAR}~\citep{tian2024visual} (scale-wise; VAR-$d16$ and VAR-$d20$). MAR takes the SAR anchor to be the decoder's representation of the class-token buffer prepended to its decoder input; VAR takes the SAR anchor to be $a = c_{\mathrm{class}}{=}\mathrm{ClassEmb}(y)$, the class embedding already consumed by AdaLN modulation. We use $T{=}256$ AR steps for MAR and $M{=}4$ equal groups at the last scale for VAR. The VAR multi-step inference is enabled by an order-agnostic random-order retraining of each VAR variant; this retraining is shared by the random and SAR rows of Table~\ref{tab:imagenet-fid} and is itself ablated in Appendix~\ref{app:var-retrain}. All other implementation details follow the original recipes and are listed in Appendix~\ref{app:impl}.

\paragraph{Evaluation.}
We evaluate on class-conditional ImageNet-1K~\citep{deng2009imagenet} at $256{\times}256$, reporting Fr\'{e}chet Inception Distance (FID)~\citep{heusel2017gans} computed against the ADM reference statistics~\citep{dhariwal2021diffusion}.

\paragraph{Results.}
Table~\ref{tab:imagenet-fid} reports class-conditional FID. SAR yields a consistent improvement over the random-order baseline at every model size and across both paradigms, despite changing only the inference-time order. Ranking tokens by similarity to the class embedding provides a more informative early context than uniform random selection.

\begin{table}[t]
\centering
\caption{Class-conditional FID ($\downarrow$) on ImageNet $256{\times}256$ across both AR paradigms. Within each column, both rows share the same checkpoint (the VAR rows use a retrained checkpoint with $M{=}4$ multi-step inference at the last scale) and the same sampling configuration; only the inference-time generation order differs.}
\label{tab:imagenet-fid}
\small
\setlength{\tabcolsep}{6pt}
\begin{tabular}{lcccc}
\toprule
 & \multicolumn{2}{c}{Patch-wise (MAR)} & \multicolumn{2}{c}{Scale-wise (VAR)} \\
 \cmidrule(lr){2-3} \cmidrule(lr){4-5}
Order  & MAR-Base & MAR-Large & VAR-$d16$ & VAR-$d20$ \\
\midrule
Random                  & 2.29          & 1.78          & 3.56          & 2.77          \\
SAR (class similarity)  & \textbf{2.21} & \textbf{1.75} & \textbf{3.48} & \textbf{2.73} \\
\bottomrule
\end{tabular}
\end{table}

\section{VAR Retraining Ablation}
\label{app:var-retrain}

Table~\ref{tab:var_sar_retrain} reports a separate training-time experiment from the inference-order comparison in Table~\ref{tab:imagenet-fid}. Here, VAR-$d16$, $d20$, $d24$, and $d30$ are retrained using SAR ordering at the last scale. The 1-step$^\ast$ row reports the original VAR setting, in which training and inference generate all tokens within each scale in one step. The remaining rows evaluate the retrained checkpoint under different inference procedures: 1-step generates all tokens within each scale at once, while random and SAR use four-step last-scale decoding with their respective position-selection rules. The 1-step results remain close to the original baseline. Under the same retrained checkpoint and four-step inference budget, SAR improves FID, IS, precision, and recall over random ordering at every model size. Thus, the random-versus-SAR comparison measures the effect of inference order after SAR-order retraining; improvements relative to the original VAR baseline reflect the combined training and inference changes and should not be attributed solely to inference-time reordering.

\begin{table}[t]
\centering
\caption{
Results of retraining VAR models with different generation orders.
For each VAR model, we apply the SAR order at the last scale during retraining.
The superscript $^\ast$ indicates the original setting where both training and evaluation generate all tokens within each scale in one step.
}
\label{tab:var_sar_retrain}
\begin{tabular}{llcccc}
\toprule
Model & Order & FID $\downarrow$ & IS $\uparrow$ & Precision $\uparrow$ & Recall $\uparrow$ \\
\midrule
\multirow{4}{*}{VAR-$d16$}
& \textcolor{gray}{1-step$^\ast$} & \textcolor{gray}{3.30} & \textcolor{gray}{274.4} & \textcolor{gray}{0.84} & \textcolor{gray}{0.51} \\
& 1-step & 3.42 & 268.5 & 0.83 & 0.49 \\
& random & 3.24 & 278.2 & 0.84 & 0.52 \\
& SAR & \textbf{3.05} & \textbf{286.2} & \textbf{0.86} & \textbf{0.54} \\
\midrule
\multirow{4}{*}{VAR-$d20$}
& \textcolor{gray}{1-step$^\ast$} & \textcolor{gray}{2.57} & \textcolor{gray}{302.6} & \textcolor{gray}{0.83} & \textcolor{gray}{0.56} \\
& 1-step & 2.68 & 297.0 & 0.82 & 0.54 \\
& random & 2.53 & 304.8 & 0.84 & 0.57 \\
& SAR & \textbf{2.35} & \textbf{314.5} & \textbf{0.85} & \textbf{0.59} \\
\midrule
\multirow{4}{*}{VAR-$d24$}
& \textcolor{gray}{1-step$^\ast$} & \textcolor{gray}{2.09} & \textcolor{gray}{312.9} & \textcolor{gray}{0.82} & \textcolor{gray}{0.59} \\
& 1-step & 2.20 & 307.4 & 0.81 & 0.57 \\
& random & 2.06 & 315.4 & 0.83 & 0.60 \\
& SAR & \textbf{1.88} & \textbf{325.3} & \textbf{0.84} & \textbf{0.62} \\
\midrule
\multirow{4}{*}{VAR-$d30$}
& \textcolor{gray}{1-step$^\ast$} & \textcolor{gray}{1.92} & \textcolor{gray}{323.1} & \textcolor{gray}{0.82} & \textcolor{gray}{0.59} \\
& 1-step & 2.03 & 317.2 & 0.81 & 0.57 \\
& random & 1.90 & 325.9 & 0.82 & 0.60 \\
& SAR & \textbf{1.74} & \textbf{335.5} & \textbf{0.84} & \textbf{0.62} \\
\bottomrule
\end{tabular}
\end{table}

\newpage
\section{Inference Latency}
\label{app:latency}
\begin{table}[!h]
\centering
\caption{VAR generation quality and end-to-end inference latency. Times are mean $\pm$ population standard deviation over $1{,}000$ images. Random and SAR share the retrained checkpoint and four-step last-scale budget.}
\label{tab:var-latency}
\small
\setlength{\tabcolsep}{4pt}
\begin{tabular}{llccccc}
\toprule
Model & Order & Time (s/image) $\downarrow$ & FID $\downarrow$ & IS $\uparrow$ & Prec. $\uparrow$ & Rec. $\uparrow$ \\
\midrule
VAR-$d16$ & 1-step & $0.073 \pm 0.002$ & 3.42 & 268.5 & 0.83 & 0.49 \\
VAR-$d16$ & random & $0.093 \pm 0.002$ & 3.24 & 278.2 & 0.84 & 0.52 \\
VAR-$d16$ & SAR & $0.093 \pm 0.002$ & \textbf{3.05} & \textbf{286.2} & \textbf{0.86} & \textbf{0.54} \\
\midrule
VAR-$d20$ & 1-step & $0.088 \pm 0.002$ & 2.68 & 297.0 & 0.82 & 0.54 \\
VAR-$d20$ & random & $0.113 \pm 0.002$ & 2.53 & 304.8 & 0.84 & 0.57 \\
VAR-$d20$ & SAR & $0.113 \pm 0.002$ & \textbf{2.35} & \textbf{314.5} & \textbf{0.85} & \textbf{0.59} \\
\midrule
VAR-$d24$ & 1-step & $0.105 \pm 0.002$ & 2.20 & 307.4 & 0.81 & 0.57 \\
VAR-$d24$ & random & $0.135 \pm 0.003$ & 2.06 & 315.4 & 0.83 & 0.60 \\
VAR-$d24$ & SAR & $0.135 \pm 0.003$ & \textbf{1.88} & \textbf{325.3} & \textbf{0.84} & \textbf{0.62} \\
\midrule
VAR-$d30$ & 1-step & $0.133 \pm 0.003$ & 2.03 & 317.2 & 0.81 & 0.57 \\
VAR-$d30$ & random & $0.175 \pm 0.004$ & 1.90 & 325.9 & 0.82 & 0.60 \\
VAR-$d30$ & SAR & $0.175 \pm 0.003$ & \textbf{1.74} & \textbf{335.5} & \textbf{0.84} & \textbf{0.62} \\
\bottomrule
\end{tabular}
\end{table}

\end{document}